\pdfoutput=1
\documentclass[11pt]{article}

\usepackage[margin=1in]{geometry}
\usepackage[T1]{fontenc}
\usepackage[utf8]{inputenc}
\usepackage{lmodern}
\usepackage{microtype}
\usepackage{amsmath}
\usepackage{amssymb}
\usepackage{graphicx}
\usepackage{booktabs}
\usepackage{tabularx}
\usepackage{array}
\usepackage{enumitem}
\usepackage{caption}
\usepackage{float}
\usepackage[section]{placeins}
\usepackage[numbers,sort&compress]{natbib}
\usepackage[colorlinks=true,linkcolor=blue,citecolor=blue,urlcolor=blue]{hyperref}

\graphicspath{{figures/}}
\setlist[itemize]{leftmargin=*,itemsep=2pt,topsep=3pt}
\setlist[enumerate]{leftmargin=*,itemsep=2pt,topsep=3pt}

\newcommand{\ewam}{EWAM}

\newcommand{\basepolicy}{Cosmos3-\allowbreak{}Nano-{}-\allowbreak{}Policy-\allowbreak{}DROID}
\newcommand{\taskbinone}{BananasIn\allowbreak{}BinOne\allowbreak{}MoreTask}
\newcommand{\taskstack}{BlockStacking\allowbreak{}OrderAgnostic\allowbreak{}Task}
\newcommand{\indicator}{\mathbb{I}}

\title{\ewam: An Enhanced World Action Model for Closed-Loop Online Adaptation in Embodied Intelligence}

\author{
Xin Zhou\textsuperscript{1} \quad Cong Miao\textsuperscript{2}\\[2pt]
\textsuperscript{1}Astronex Robotics\\
\textsuperscript{2}Nanjing University of Information Science and Technology\\[2pt]
}

\date{}

\begin{document}
\maketitle

\begin{abstract}
Robotic deployment in open environments remains constrained by the cost of collecting demonstrations that cover diverse object poses, scene layouts, and execution contingencies. World Action Models (WAMs) couple future-state prediction with action generation, but static offline pipelines remain vulnerable to execution-level mismatch such as collisions, empty grasps, and perception-induced hallucinations. We introduce the Enhanced World Action Model (\ewam), a closed-loop online adaptation architecture built upon a pretrained frozen \basepolicy{} backbone and evaluated under a zero-shot task protocol. \ewam\ adds four trainable layers to the frozen policy path: a Neural Experience Memory Layer inside the Diffusion Transformer (DiT), a Neural Anomaly Detection Layer after the state prediction head, a Neural Policy Routing Layer after anomaly detection, and a Neural Action Correction Layer after the action output head. Memory retrieval supplies task-relevant execution context, anomaly detection monitors prediction-realization divergence, routing selects direct execution, conservative replanning, or rollback recovery, and action correction refines generated action chunks using execution diagnostics. On the BananaInBowlTask local test, \ewam\ matches the 100\% task success rate of \basepolicy{} while reducing completion time from 25.60 s to 9.27 s, path length from 1.81 m to 0.83 m, and total execution faults from 13.5 to 2.2 per episode.
\end{abstract}


\section{Introduction}

World models have long been used to learn compact predictive representations of environment dynamics \citep{ha2018world,hafner2020dreamer,dreamerv3_2023}. In embodied intelligence, this idea has recently evolved into World Action Models (WAMs), where future-state prediction is coupled directly with robot action generation. DreamZero studies WAMs as zero-shot policies, Fast-WAM studies whether expensive test-time imagination is necessary, GigaWorld-Policy studies efficient action-centered WAM design, and Cosmos 3 provides an omnimodal foundation for physical AI \citep{dreamzero2026,fastwam2026,gigaworldpolicy2026,cosmos3_2026}.

\textbf{Backbone and comparison rationale.} \ewam\ uses a frozen \basepolicy{} WAM backbone as the implementation base \citep{cosmos3_2026}. The experimental question is deployment-oriented: given the same policy interface, can closed-loop layers improve execution quality on contact-rich RoboLab tasks? The reported evidence is limited to the RoboLab manipulation slice summarized in Table~\ref{tab:multi_task}.

From a deployment perspective, however, existing WAM pipelines still face three practical bottlenecks when moved to new layouts under limited additional data. First, collecting offline demonstrations and annotated trajectories for every long-tail object pose or scene state remains expensive. Second, many failures arise from execution-level divergence between predicted and realized states rather than from semantic misunderstanding. Typical examples include container-edge collision, empty grasp, and high-confidence perception with low physical contact. Third, once deployed, static WAMs do not efficiently convert successful local experience into future action advantages. Unfiltered online learning can degrade policy quality when unsafe or redundant trajectories are written back into the training stream.

\ewam\ starts from the \basepolicy{} initialization and keeps the backbone frozen. In the reported zero-shot protocol, no new demonstration set is collected for the target RoboLab tasks. Deployment-time adaptation is handled by four inserted neural layers: a Neural Experience Memory Layer for experience-augmented DiT context, a Neural Anomaly Detection Layer for execution mismatch detection, a Neural Policy Routing Layer for strategy selection, and a Neural Action Correction Layer for action refinement. A rollback manager recovers from unstable execution states, and an experience filter blocks low-quality trajectories before memory writing or online parameter updates. The method targets five execution problems: prediction-realization mismatch, collision, empty grasp, perception hallucination, and trajectory redundancy.

Our contributions are:
\begin{itemize}
    \item We formulate \ewam\ as a \basepolicy{}-based architecture with four inserted neural layers: Neural Experience Memory Layer at DiT intermediate layers, Neural Anomaly Detection Layer after state prediction, Neural Policy Routing Layer after anomaly detection, and Neural Action Correction Layer after action output, forming a closed-loop inference path with trainable differentiable modules and supervised discrete routing.
    \item We introduce a branch-aware experience filtering and rollback design that prevents unsafe or low-value trajectories from contaminating memory and online updates.
    \item We report simulation results and ablations on RoboLab zero-shot manipulation tasks, showing improved execution quality while preserving the frozen \basepolicy{} backbone.
\end{itemize}

\section{Related Work}

\paragraph{World models and World Action Models.}
World Models, Dreamer, and DreamerV3 established the value of learning predictive latent dynamics for decision making \citep{ha2018world,hafner2020dreamer,dreamerv3_2023}. Recent robot learning work extends this idea toward WAMs that connect world prediction with action generation. DreamDojo studies large-scale human-video world modeling, DreamZero studies zero-shot WAM policies, Fast-WAM studies efficient WAM inference, GigaWorld-Policy studies action-centered WAM design, LingBot-VA studies causal world modeling for robot control, and Motus unifies latent action representations \citep{dreamdojo2026,dreamzero2026,fastwam2026,gigaworldpolicy2026,causal_world_modeling2026,motus2025}. \ewam\ follows the lightweight deployment-oriented branch of this literature: instead of adding a heavier imagination module, it focuses on retrievable experience, detectable execution error, rollback recovery, and incremental adaptation.

\paragraph{Vision-language-action models.}
SayCan and RT-1 grounded language-conditioned robot execution at scale, while RT-2, OpenVLA, $\pi_0/\pi_{0.5}$, ABot-M0, and GR-2 study semantic transfer through vision-language-action pretraining \citep{saycan2023,rt1_2023,rt2_2023,openvla2024,pi0_2024,pi05_2025,abotm0_2026,gr2_2024}. VLA models focus on language-visual semantic understanding and mapping natural-language instructions to actions. In contrast, WAM models, including Cosmos3 and \ewam, emphasize world-state prediction, physical consistency, and contact-sensitive action generation. Thus, VLA models and \ewam\ are complementary rather than interchangeable.

\paragraph{Action diffusion, asynchronous control, and feedback.}
Diffusion Policy, planning with diffusion, and Diffusion Transformers show that diffusion-style action generation can represent multimodal robot behavior and scale with transformer backbones \citep{diffusion_policy2023,planning_diffusion2022,dit2023}. UniPi and Video Prediction Policy further connect video prediction with policy learning \citep{unipi2023,video_prediction_policy2025}. \ewam\ shares the deployment concern of efficient feedback, but ties memory retrieval to the DiT reasoning path and ties filtering, rollback, and correction to execution diagnostics.

\paragraph{Simulation benchmarks.}
RoboLab provides high-fidelity simulation for analyzing task-generalist policies through visual, procedural, and relational evaluation dimensions \citep{robolab2026}. LIBERO supports lifelong robot learning evaluation, while RoboMimic provides a widely used offline manipulation benchmark for comparing imitation-learning pipelines and dataset-quality effects \citep{libero2023,robomimic2021}. This paper evaluates \ewam\ on three zero-shot manipulation tasks in RoboLab.

\subsection{Comparison Models}

Table~\ref{tab:related_comparison} positions \ewam\ against the requested VLA and WAM baselines. The table is capability-level because the models are not all reported under the same task, action space, and robot morphology. Measured numerical claims for \ewam\ in this paper are restricted to the local RoboLab protocol in Tables~\ref{tab:local_banana_comparison}--\ref{tab:multi_task}.

\begin{table}[!htbp]
\centering
\caption{Capability-level comparison with the requested baselines.}
\label{tab:related_comparison}
\small
\begin{tabularx}{\textwidth}{p{0.18\textwidth}p{0.15\textwidth}p{0.32\textwidth}X}
\toprule
Model & Family & Main modeling emphasis & Relation to \ewam \\
\midrule
$\pi_0$ \citep{pi0_2024} & VLA & Flow-based continuous action generation & Flow-policy VLA baseline; lacks explicit WAM anomaly routing. \\
$\pi_{0.5}$ \citep{pi05_2025} & VLA & Open-world VLA generalization & Semantic generalization baseline. \\
ABot-M0 \citep{abotm0_2026} & VLA & Action manifold learning with DiT action prediction & Efficient action manifold baseline. \\
Motus from Wan-family backbone \citep{motus2025,wan2025} & WAM / video-action & Motus-style latent action modeling initialized from a Wan video backbone & Video-foundation WAM variant. \\
Motus \citep{motus2025} & WAM & Unified latent action world model & Latent-action WAM baseline. \\
LingBot-VA \citep{causal_world_modeling2026} & WAM / VLA & Causal world modeling with video-action rollout & Causal rollout baseline. \\
Fast-WAM \citep{fastwam2026} & WAM & Efficient action generation without heavy test-time imagination & Fast WAM baseline. \\
\textbf{\ewam} & \textbf{WAM adaptation} & \textbf{Frozen \basepolicy{} backbone with memory, anomaly, routing, and correction layers} & \textbf{Closed-loop online adaptation.} \\
\bottomrule
\end{tabularx}
\end{table}

The fundamental distinction is that \ewam\ treats online adaptation as an \textbf{inference-layer closed-loop problem}: the backbone remains frozen, while inserted neural layers provide memory retrieval, anomaly detection, policy routing, and action correction.

\section{Method}

\subsection{Problem Formulation}

At time step $t$, the robot receives a visual observation $o_t$, proprioceptive state $q_t$, and language instruction $l$. \ewam\ is built upon the \basepolicy{} backbone, which consists of a Vision-Language Encoder (VLE), an Autoregressive (AR) Reasoner, and a Diffusion Transformer (DiT) action generator. We insert four neural layers into this backbone: (1) Neural Experience Memory Layer at DiT intermediate layers, (2) Neural Anomaly Detection Layer after the state prediction head, (3) Neural Policy Routing Layer after anomaly detection, and (4) Neural Action Correction Layer after the action output head.

The core forward pass with inserted neural layers is:
\begin{align}
h_{\mathrm{VLE}} &= \mathrm{VLE}_\phi(o_t, q_t, l), \quad \text{vision-language encoding} \\
c_t &= \mathrm{ARReasoner}_\phi(h_{\mathrm{VLE}}), \quad \text{autoregressive reasoning context} \\
h_{\mathrm{DiT}}^{(l)} &= \mathrm{DiTLayer}_\phi^{(l)}(h_{\mathrm{DiT}}^{(l-1)}), \quad l < l_{\mathrm{mem}} \\
h_{\mathrm{DiT}}^{(l_{\mathrm{mem}})} &= \mathrm{NeuralMemoryLayer}_\theta(\mathrm{DiTLayer}_\phi^{(l_{\mathrm{mem}})}(h_{\mathrm{DiT}}^{(l_{\mathrm{mem}}-1)}), M, c_t), \quad \text{memory-augmented DiT layer} \\
h_{\mathrm{DiT}}^{(L)} &= \mathrm{DiTLayer}_\phi^{(L)}(h_{\mathrm{DiT}}^{(L-1)}), \quad l_{\mathrm{mem}} < l \le L \\
\hat{s}_{t+1} &= \mathrm{StateHead}_\phi(h_{\mathrm{DiT}}^{(L)}), \quad \text{state prediction} \\
a_{t:t+H}^{0} &= \mathrm{ActionHead}_\phi(h_{\mathrm{DiT}}^{(L)}), \quad \text{raw action output} \\
\iota_t &= \mathrm{NeuralAnomalyLayer}_\theta(\hat{s}_{t+1}, s_t, a_{t:t+H}^{0}, h_{\mathrm{DiT}}^{(L)}), \quad \text{anomaly detection vector} \\
r_t &= \mathrm{NeuralRoutingLayer}_\theta(\iota_t, c_t, M), \quad \text{policy routing decision} \\
a_{t:t+H}^\ast &= \mathrm{NeuralCorrectionLayer}_\theta(a_{t:t+H}^{0}, \iota_t, r_t, M), \quad \text{corrected action}
\end{align}
where:
\begin{itemize}
    \item $h_{\mathrm{VLE}}$: Vision-language encoder hidden state
    \item $c_t$: Autoregressive reasoning context vector
    \item $h_{\mathrm{DiT}}^{(l)}$: Hidden state at DiT layer $l$ (the superscript $(l)$ denotes layer index)
    \item $l_{\mathrm{mem}}=6$: Insertion layer for Neural Experience Memory Layer
    \item $M$: Experience memory database
    \item $\hat{s}_{t+1}$: Predicted next world state from state prediction head
    \item $\iota_t \in \mathbb{R}^6$: Anomaly detection vector containing previous prediction residual, dynamics inconsistency, collision score, empty-grasp score, hallucination score, and force-violation score
    \item $r_t \in \{\mathrm{direct}, \mathrm{conservative}, \mathrm{rollback}\}$: Policy routing decision
    \item $a_{t:t+H}^{0}$: Raw robot action chunk in the \basepolicy{} action-token interface
    \item $a_{t:t+H}^\ast$: Final corrected action chunk after Neural Action Correction Layer
\end{itemize}
The \basepolicy{} backbone parameters $\phi$ are frozen; only the four neural layer parameters $\theta$ and lightweight adapters are trained or updated. In this paper, zero-shot refers to the target-task evaluation protocol: the target RoboLab tasks provide no additional demonstration dataset, and adaptation comes from memory retrieval, filtered online experience, and lightweight updates.

\subsection{Overall Architecture}

\begin{figure}[H]
    \centering
    \includegraphics[width=0.88\textwidth]{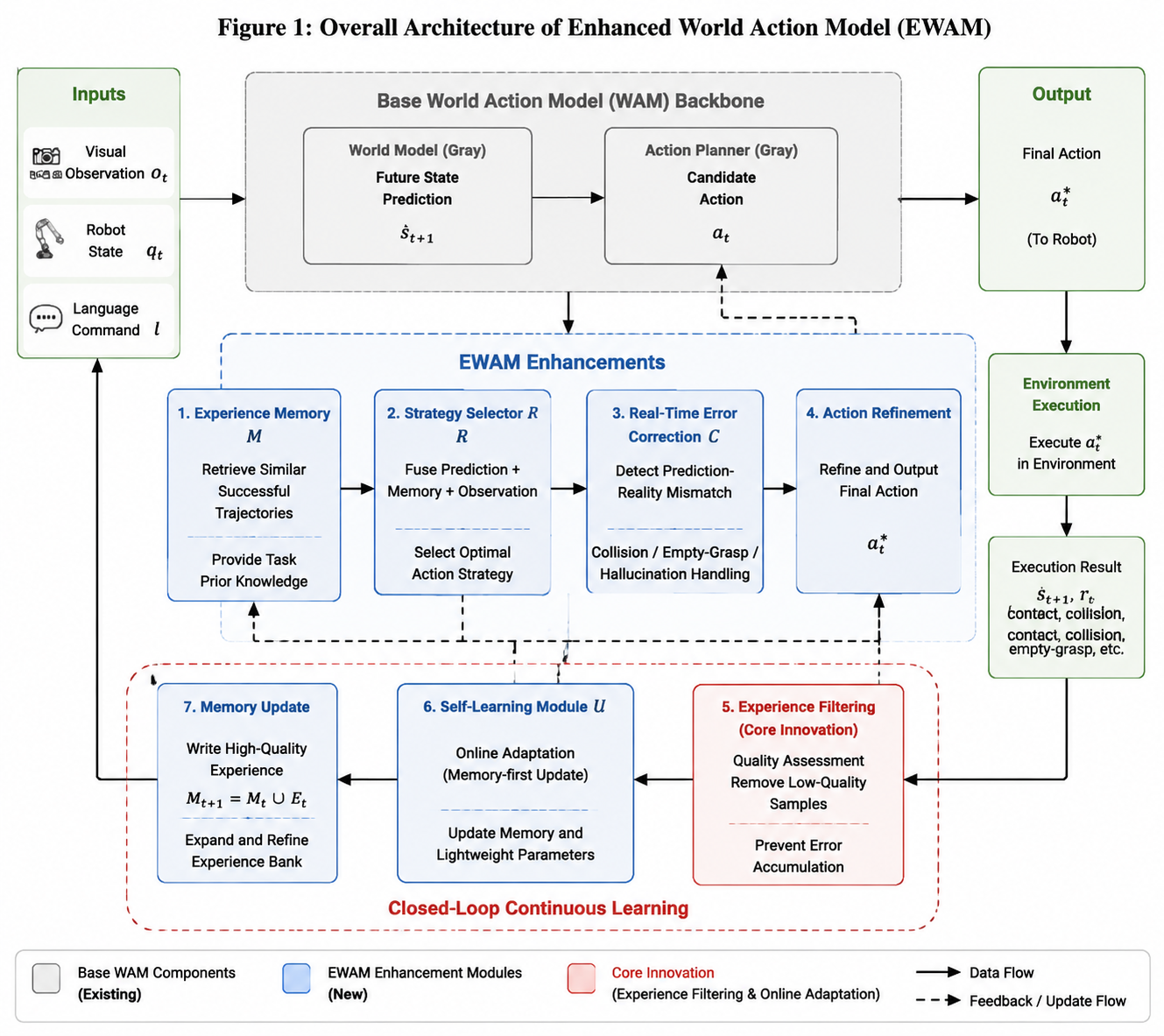}
    \caption{Overall architecture of \ewam\ built upon \basepolicy{}. Four neural layers are inserted: Neural Experience Memory Layer at DiT intermediate layers, Neural Anomaly Detection Layer after state prediction, Neural Policy Routing Layer after anomaly detection, and Neural Action Correction Layer after action output.}
    \label{fig:architecture}
\end{figure}

Figure~\ref{fig:architecture} summarizes the \ewam\ architecture built upon \basepolicy{}. The backbone consists of three main components: (1) Vision-Language Encoder (VLE) for multimodal perception, (2) Autoregressive (AR) Reasoner for long-horizon task planning, and (3) Diffusion Transformer (DiT) for action generation. We insert four neural layers to enable closed-loop online adaptation:

\textbf{Neural Experience Memory Layer (at DiT intermediate layers):} Inserted at layer $l_{\mathrm{mem}}$ of the DiT, this layer retrieves task-relevant experiences from memory $M$ and injects them into the DiT hidden representations. It enables the diffusion process to be conditioned on both the current context and retrieved historical experience.

\textbf{Neural Anomaly Detection Layer (after state prediction head):} This layer takes the predicted state $\hat{s}_{t+1}$, current state $s_t$, raw action candidate $a_{t:t+H}^{0}$, and DiT hidden state to detect execution anomalies including prediction-realization mismatch, collision risk, empty grasp, perception hallucination, and force violation.

\textbf{Neural Policy Routing Layer (after anomaly detection):} Based on the anomaly detection results $\iota_t$, this layer selects the appropriate execution strategy: direct execution, conservative replanning, or rollback recovery. It outputs a routing decision $r_t$ that modulates the action correction process.

\textbf{Neural Action Correction Layer (after action output head):} This layer refines the raw action output $a_{t:t+H}^{0}$ based on anomaly signals $\iota_t$, routing decision $r_t$, and memory context $M$. It performs denoising-time correction to ensure safe and efficient execution.

The \basepolicy{} backbone parameters $\phi$ remain frozen; only the four neural layer parameters $\theta$ and lightweight adapters are trained, enabling efficient adaptation while preserving the pretrained knowledge.

\begin{table}[!htbp]
\centering
\caption{Key \ewam\ configuration.}
\label{tab:hyperparameters}
\small
\begin{tabularx}{\textwidth}{@{}p{0.24\textwidth}X@{}}
\toprule
Component & Configuration \\
\midrule
Memory insertion & DiT layer $l_{\mathrm{mem}}=6$; top-$k=5$ retrieval; memory gate $\alpha_{\mathrm{mem}}=0.4$. \\
Anomaly thresholds & Base threshold $\tau_0=0.65$; grasp confidence threshold $g_{\mathrm{hi}}=0.85$; low-force threshold $f_{\mathrm{lo}}=0.15$ N. \\
Routing thresholds & Direct execution for anomaly score $<0.3$; conservative replanning for $[0.3,0.7)$; rollback for $\ge0.7$. \\
Correction layer & Hidden dimension 64; correction strength $\alpha_{\mathrm{corr}}=0.5$; maximum joint correction 0.3 rad. \\
Training weights & Action/state/memory/anomaly/routing/correction weights: $1.0/0.5/0.3/0.8/0.6/1.2$. \\
Online update & Learning rate $10^{-4}$; adapter rank 16; replay buffer size 1000; updates applied only after quality filtering. \\
Action interface & \basepolicy{} robot action tokens: camera pose delta, end-effector pose delta, and gripper state; local rollouts pack these tokens into fixed-horizon chunks. \\
\bottomrule
\end{tabularx}
\vspace{2pt}
\footnotesize The full \basepolicy{} backbone is frozen; only inserted layers and lightweight adapters are updated.
\end{table}

\subsection{Neural Experience Memory Layer}

The Neural Experience Memory Layer is inserted at the intermediate layer $l_{\mathrm{mem}}$ of the DiT. It serves as a bridge between the diffusion process and the experience memory $M$, enabling the action generation to be conditioned on both current context and retrieved historical experience.

The layer operates as follows:
\begin{align}
q_{\mathrm{mem}} &= W_q h_{\mathrm{DiT}}^{(l_{\mathrm{mem}}-1)}, \quad \text{query from DiT hidden state} \\
k_m, v_m &= \mathrm{MemoryEncoder}(M, c_t), \quad \text{encode memory with AR context} \\
\mathrm{Attn}_{\mathrm{mem}} &= \mathrm{Softmax}\left(\frac{q_{\mathrm{mem}} k_m^\top}{\sqrt{d_k}}\right) v_m, \quad \text{memory attention} \\
h_{\mathrm{DiT}}^{(l_{\mathrm{mem}})} &= h_{\mathrm{DiT}}^{(l_{\mathrm{mem}}-1)} + \alpha_{\mathrm{mem}} \cdot \mathrm{Attn}_{\mathrm{mem}}, \quad \text{residual connection}
\end{align}
where $W_q$ is a learnable query projection, $\mathrm{MemoryEncoder}$ encodes the experience memory conditioned on the AR reasoning context $c_t$, and $\alpha_{\mathrm{mem}}$ is a gating scalar. This design allows the DiT to attend to relevant past experiences during action generation, providing task-relevant priors and rollback anchors.

\subsection{Neural Anomaly Detection Layer}

The Neural Anomaly Detection Layer is inserted after state and raw-action prediction. At decision time, it combines the previous prediction residual, candidate-action dynamics consistency, and learned risk scores. After the corrected action is executed, the realized transition residual is written into the experience record and used for filtering and online updates.

The anomaly detection is formulated as:
\begin{align}
e_{\mathrm{prev}} &= \|s_t - \hat{s}_t\|_2, \quad \text{previous prediction residual} \\
e_{\mathrm{dyn}} &= \|\hat{s}_{t+1} - \mathrm{DynamicsModel}(s_t, a_{t:t+H}^{0})\|_2, \quad \text{candidate dynamics inconsistency} \\
\iota_{\mathrm{collision}} &= \sigma(W_c h_{\mathrm{DiT}}^{(L)} + b_c), \quad \text{collision probability} \\
\iota_{\mathrm{grasp}} &= \sigma(W_g h_{\mathrm{DiT}}^{(L)} + b_g), \quad \text{empty grasp probability} \\
\iota_{\mathrm{halluc}} &= \sigma(W_h h_{\mathrm{DiT}}^{(L)} + b_h), \quad \text{hallucination probability} \\
\iota_{\mathrm{force}} &= \sigma(W_f h_{\mathrm{DiT}}^{(L)} + b_f), \quad \text{force violation probability} \\
\iota_t &= (e_{\mathrm{prev}}, e_{\mathrm{dyn}}, \iota_{\mathrm{collision}}, \iota_{\mathrm{grasp}}, \iota_{\mathrm{halluc}}, \iota_{\mathrm{force}}),
\end{align}
where $\sigma$ is the sigmoid function, and $W_c, W_g, W_h, W_f$ are learnable weights. At the first step, $e_{\mathrm{prev}}$ is initialized to zero. The anomaly vector $\iota_t$ provides execution diagnostics for downstream routing and correction. The realized residual $e_t=\|s_{t+1}-\hat{s}_{t+1}\|_2$ is computed after environment execution and is used by the rollback and filtering logic in Section~\ref{sec:error_rollback}.

\subsection{Neural Policy Routing Layer}

The Neural Policy Routing Layer is inserted after the anomaly detection layer. It selects the appropriate execution strategy based on the anomaly signals $\iota_t$, AR context $c_t$, and memory $M$.

The routing decision is computed as:
\begin{align}
h_{\mathrm{route}} &= \mathrm{MLP}([\iota_t; c_t; \mathrm{MemSummary}(M)]), \\
p_{\mathrm{route}} &= \mathrm{Softmax}(W_{\mathrm{route}} h_{\mathrm{route}}), \\
r_t &= \arg\max(p_{\mathrm{route}}) \in \{\mathrm{direct}, \mathrm{conservative}, \mathrm{rollback}\},
\end{align}
where $[\cdot;\cdot]$ denotes concatenation, $\mathrm{MemSummary}(M)$ provides a compact summary of the memory state, and $p_{\mathrm{route}}$ is a probability distribution over three strategies:
\begin{itemize}
    \item \textbf{Direct execution}: Proceed with the generated action when anomaly signals are below threshold.
    \item \textbf{Conservative replanning}: Trigger action correction with conservative constraints when mild anomalies are detected.
    \item \textbf{Rollback recovery}: Restore to the nearest stable state from memory when severe anomalies are detected.
\end{itemize}

\subsection{Neural Action Correction Layer}

The Neural Action Correction Layer is inserted after the action output head. It refines the raw action output $a_{t:t+H}^{0}$ based on anomaly signals $\iota_t$, routing decision $r_t$, and memory context $M$.

The correction process depends on the routing decision:
\begin{align}
a_{t:t+H}^\ast &=
\begin{cases}
a_{t:t+H}^{0}, & r_t = \mathrm{direct} \\
a_{t:t+H}^{0} + \alpha_{\mathrm{corr}} \cdot \mathrm{CorrectionNet}(a_{t:t+H}^{0}, \iota_t, M), & r_t = \mathrm{conservative} \\
\mathrm{RollbackAction}(M, \iota_t), & r_t = \mathrm{rollback}
\end{cases}
\end{align}
where $\mathrm{CorrectionNet}$ is a lightweight neural network that predicts action corrections conditioned on anomaly signals and memory, $\alpha_{\mathrm{corr}}$ is a correction strength scalar, and $\mathrm{RollbackAction}$ retrieves a conservative action from the nearest stable state in memory.

For conservative replanning, the correction network is trained to minimize:
\begin{equation}
L_{\mathrm{corr}} = \indicator[\mathrm{anomaly}]\|a_{t:t+H}^\ast - a_{t:t+H}^{\mathrm{safe}}\|_2^2,
\end{equation}
where $a_{t:t+H}^{\mathrm{safe}}$ is a safe action obtained from simulator rollback, conservative recovery primitives, or a previously admitted successful memory item. The reported zero-shot experiments do not add a new human demonstration set for the target tasks; they test whether the inserted \ewam\ layers improve task adaptation by combining the pretrained WAM prior with filtered online feedback.

\subsection{Closed-Loop Online Learning}

\begin{figure}[H]
    \centering
    \includegraphics[width=0.86\textwidth]{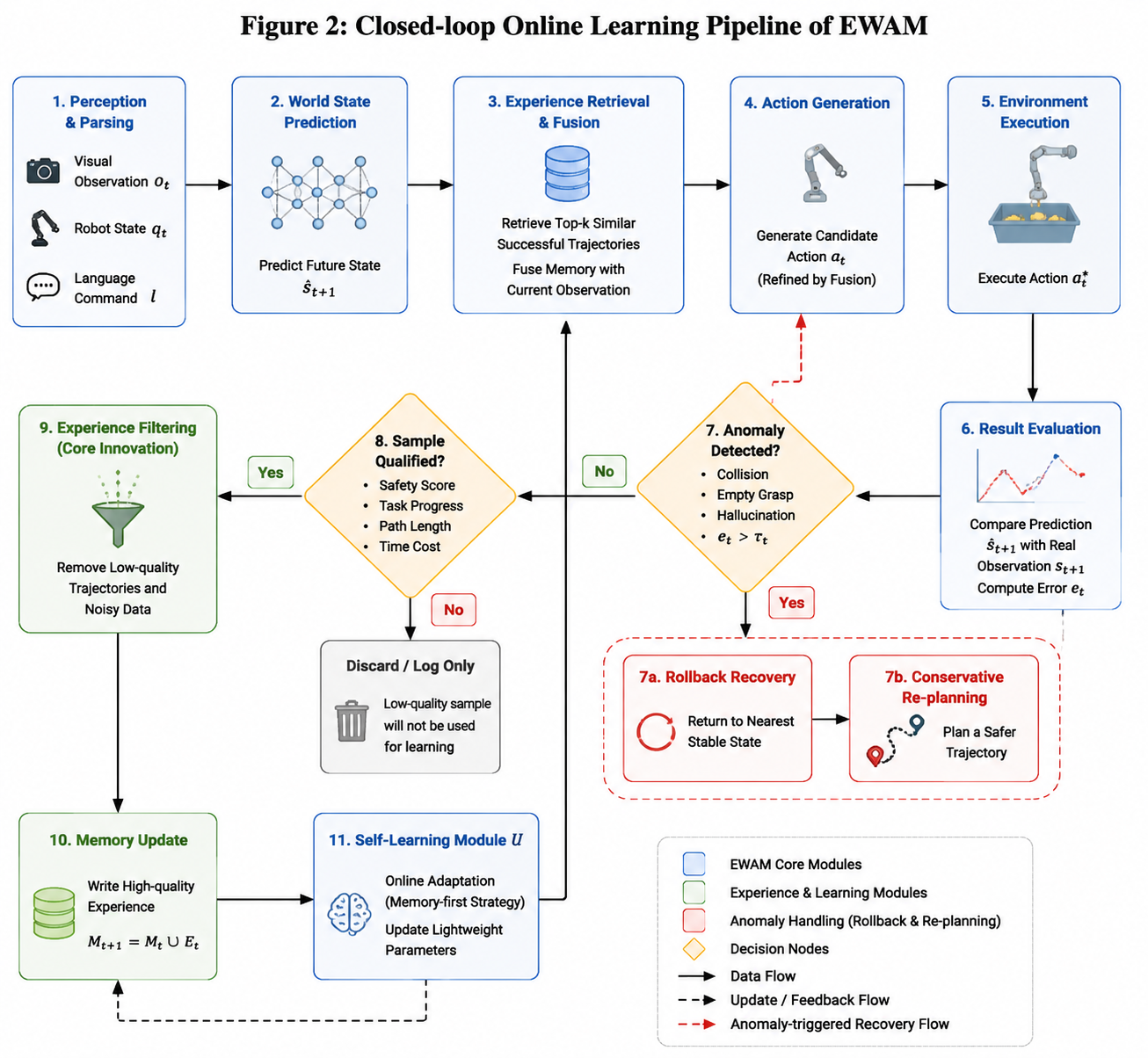}
    \caption{Closed-loop online learning pipeline of \ewam. The system routes successful high-quality trajectories into memory and lightweight updates, while anomalies trigger rollback and conservative replanning.}
    \label{fig:closedloop}
\end{figure}

The online learning pipeline is shown in Figure~\ref{fig:closedloop}. \ewam\ operates as a closed-loop cycle with the following data flow:

\begin{enumerate}
    \item \textbf{Perception $\rightarrow$ VLE encoding}: Raw visual observation $o_t$ and proprioceptive state $q_t$ are encoded by the Vision-Language Encoder to produce $h_{\mathrm{VLE}}$.
    \item \textbf{Encoding $\rightarrow$ AR reasoning}: The autoregressive reasoner processes $h_{\mathrm{VLE}}$ to produce reasoning context $c_t$.
    \item \textbf{Reasoning $\rightarrow$ DiT with memory augmentation}: The DiT processes autoregressive tokens with memory-augmented intermediate layers (Neural Experience Memory Layer at $l_{\mathrm{mem}}=6$) that retrieve relevant experiences from $M$.
    \item \textbf{DiT output $\rightarrow$ State/action candidates}: The state prediction head produces $\hat{s}_{t+1}$ and the action head produces raw action chunk $a_{t:t+H}^{0}$.
    \item \textbf{State/action candidates $\rightarrow$ Anomaly detection}: The Neural Anomaly Detection Layer combines previous prediction residuals, candidate dynamics consistency, and learned risk scores to produce anomaly vector $\iota_t$.
    \item \textbf{Anomaly detection $\rightarrow$ Policy routing}: The Neural Policy Routing Layer selects execution strategy $r_t \in \{\mathrm{direct}, \mathrm{conservative}, \mathrm{rollback}\}$ based on $\iota_t$.
    \item \textbf{Routing + DiT output $\rightarrow$ Action correction}: The Neural Action Correction Layer refines raw action $a_{t:t+H}^0$ to corrected action $a_{t:t+H}^*$ based on $\iota_t$, $r_t$, and memory context.
    \item \textbf{Corrected action $\rightarrow$ Environment execution}: The corrected action is executed in the environment.
    \item \textbf{Execution result $\rightarrow$ Experience filtering}: The execution outcome is evaluated by the experience filter; only qualified trajectories proceed.
    \item \textbf{Filtered experience $\rightarrow$ Memory/parameter update}: Qualified trajectories are written to experience memory $M$; intermittent parameter updates are applied to neural layers.
\end{enumerate}

The loop triggers conservative replanning when the Neural Anomaly Detection Layer detects prediction error, collision risk, empty grasp, or hallucination, and triggers rollback recovery when severe anomalies are detected. Only qualified trajectories are written to memory and used for neural layer parameter updates.

\textbf{Design rationale: memory-first, adapter-second for zero-shot evaluation.} Under limited additional task data, directly updating the pretrained WAM backbone with sparse online experience risks catastrophic forgetting, where the model loses its general capabilities learned from the original dataset. \ewam\ therefore adopts a conservative adaptation strategy:
\begin{enumerate}
    \item \textbf{Immediate memory retrieval}: When a new task begins, the system first retrieves relevant experiences from memory ($k=5$ nearest neighbors) to provide task-specific context. This retrieval-based adaptation is instantaneous and does not modify any model parameters.
    \item \textbf{Conservative parameter updates}: Online parameter updates occur only intermittently and affect only the lightweight neural layer adapters, not the frozen \basepolicy{} backbone. This preserves the model's general zero-shot capabilities while allowing gradual adaptation.
    \item \textbf{Experience filtering}: Only high-quality trajectories (passing all safety and efficiency checks) are written to memory. This prevents erroneous behaviors from contaminating the experience database.
\end{enumerate}
The ablation results support this design: memory-only (12.67s) outperforms update-only (14.67s), indicating that retrieval-based adaptation is more effective in this setting than immediate parameter updates from sparse online data.

\subsection{Training and Online Update Objectives}

Offline preparation follows the base WAM objective and adds supervised losses for the four neural layers when simulator labels or admitted recovery targets are available:
\begin{align}
L_{\mathrm{state}} &= \|\hat{s}_{t+1}-s_{t+1}\|_2^2,\\
L_{\mathrm{act}} &= \|a_{t:t+H}^{0}-a_{t:t+H}^{\mathrm{gt}}\|_2^2,\\
L_{\mathrm{mem}} &= 1-\cos(\psi(o_t,l),\psi(m_t)),\\
L_{\mathrm{anomaly}} &= \mathrm{BCE}(\rho_t, \rho_t^{\mathrm{gt}})+\|(e_{\mathrm{prev}},e_{\mathrm{dyn}})-(e_{\mathrm{prev}}^{\mathrm{gt}},e_{\mathrm{dyn}}^{\mathrm{gt}})\|_2^2,\\
L_{\mathrm{route}} &= \mathrm{CE}(r_t, r_t^{\mathrm{gt}}),\\
L_{\mathrm{corr}} &= \indicator[\mathrm{anomaly}]\|a_{t:t+H}^\ast - a_{t:t+H}^{\mathrm{safe}}\|_2^2,
\end{align}
where $a_{t:t+H}^{\mathrm{gt}}$ is the reference action chunk when available, $\rho_t=(\iota_{\mathrm{collision}},\iota_{\mathrm{grasp}},\iota_{\mathrm{halluc}},\iota_{\mathrm{force}})$ denotes the binary risk heads, $\rho_t^{\mathrm{gt}}$ is the simulator-derived risk label, $r_t^{\mathrm{gt}}$ is the routing label induced by the recovery policy, and BCE and CE denote binary cross-entropy and cross-entropy losses. The combined objective is
\begin{equation}
L=\lambda_aL_{\mathrm{act}}+\lambda_sL_{\mathrm{state}}+\lambda_mL_{\mathrm{mem}}+\lambda_{\iota}L_{\mathrm{anomaly}}+\lambda_rL_{\mathrm{route}}+\lambda_cL_{\mathrm{corr}},
\end{equation}
where $(\lambda_a,\lambda_s,\lambda_m,\lambda_{\iota},\lambda_r,\lambda_c)=(1.0,0.5,0.3,0.8,0.6,1.2)$, matching the reported configuration in Table~\ref{tab:hyperparameters}. The pretrained \basepolicy{} backbone is frozen; only the four neural layer parameters $\theta$ and lightweight adapters are trained. During deployment, \ewam\ updates memory every time a sample passes filtering and updates neural layer parameters only intermittently:
\begin{align}
M_{t+1} &= M_t \oplus \Delta E_t,\\
\theta_{t+1} &= \theta_t-\eta\nabla_\theta L_{\mathrm{online}}.
\end{align}

\subsection{Error Triggering and Rollback}
\label{sec:error_rollback}

The basic prediction error is
\begin{equation}
e_t=\|s_{t+1}-\hat{s}_{t+1}\|_2.
\end{equation}
We use a dynamic threshold:
\begin{equation}
\tau_t=\tau_0+\alpha\sigma_{e,t}+\beta \indicator_{\mathrm{contact},t}+\gamma \indicator_{\mathrm{hall},t},
\end{equation}
where $\sigma_{e,t}$ is recent error-window variance, $\indicator_{\mathrm{contact},t}$ indicates abnormal contact, and $\indicator_{\mathrm{hall},t}$ indicates perception hallucination or empty grasp. Correction is triggered when
\begin{equation}
e_t>\tau_t \quad \mathrm{or} \quad \mathrm{collision}_t=1 \quad \mathrm{or} \quad \mathrm{empty\_grasp}_t=1 \quad \mathrm{or} \quad \mathrm{hallucination}_t=1.
\end{equation}
A high-confidence grasp with low physical contact is treated as hallucinated execution:
\begin{equation}
\indicator_{\mathrm{hall},t}=\indicator[g_t>g_{\mathrm{hi}}\land f_t<f_{\mathrm{lo}}],
\end{equation}
where $g_{\mathrm{hi}}=0.85$ and $f_{\mathrm{lo}}=0.15$ N in the RoboLab calibration.

The online rollout can be expressed as a gated dynamical system:
\begin{align}
(\hat{s}_{t+1},a_t) &= W_\phi(o_t,q_t,l),\\
m_t &= \mathrm{Retrieve}_k(M_t,o_t,q_t,l),\\
z_t &= \mathrm{Fuse}(o_t,q_t,\hat{s}_{t+1},m_t),\\
\tilde{a}_t &= \mathrm{Refine}_\theta(z_t,a_t),\\
\delta_t &= \indicator[e_t>\tau_t \lor \mathrm{collision}_t \lor \mathrm{empty\_grasp}_t \lor \mathrm{hallucination}_t],\\
s_t^{\mathrm{rb}} &= \mathrm{NearestStable}(B_t),\\
a_t^\ast &= (1-\delta_t)\tilde{a}_t+\delta_t\,\mathrm{Replan}(s_t^{\mathrm{rb}},M_t;\mathrm{conservative}),\\
(s_{t+1},\iota_t) &= \mathcal{E}(s_t,a_t^\ast),\\
E_t &= \mathrm{Pack}(o_t,q_t,a_t^\ast,\hat{s}_{t+1},s_{t+1},\iota_t),
\end{align}
where $\mathcal{E}$ denotes the RoboLab transition operator and $\iota_t$ contains execution diagnostics. Memory and lightweight updates are then controlled by the quality gate $Q(E_t)$:
\begin{align}
M_{t+1} &= M_t \cup \{E_t:Q(E_t)=1\},\\
B_{t+1} &= \mathrm{UpdateStableBuffer}(B_t,E_t,Q(E_t)),\\
\theta_{t+1} &=
\begin{cases}
\theta_t-\eta\nabla_\theta L_{\mathrm{online}}(\theta_t;\mathrm{Replay}(M_{t+1})), & Q(E_t)=1 \land U_t=1,\\
\theta_t, & \mathrm{otherwise},
\end{cases}
\end{align}
where $U_t$ is a scheduled-update indicator.

\subsection{Experience Filtering}

Unfiltered online learning can absorb unsafe trajectories and update the policy in the wrong direction. \ewam\ therefore applies a quality gate before memory writing or parameter updates. A sample is treated as qualified only if all conditions hold:
\begin{equation}
\begin{cases}
\mathrm{SafetyScore} \ge 0.75,\\
\mathrm{TaskScore} \ge 0.80,\\
\mathrm{GripperForce}_{\max} \le 0.80,\\
\mathrm{EE\ SPARC} \le -2.5,\\
\mathrm{PathLen} \le 1.2\ \mathrm{m},\\
\mathrm{TaskTime} \le 15\ \mathrm{s}.
\end{cases}
\end{equation}
Low-quality samples are logged for diagnostics but are not written into memory and do not contribute to online updates.

\begin{figure}[H]
    \centering
    \includegraphics[width=0.82\textwidth]{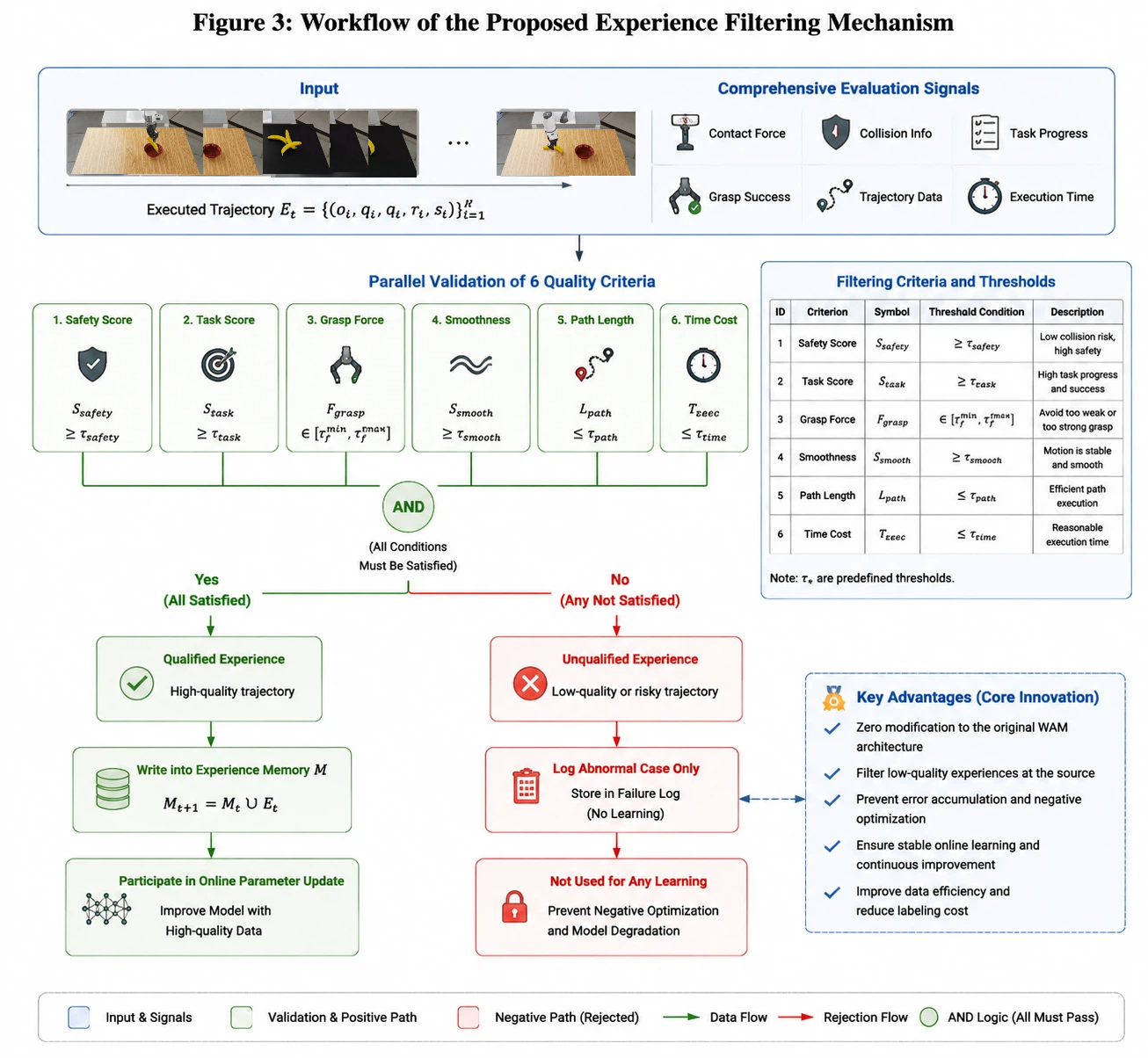}
    \caption{Experience filtering and memory-admission logic. The quality gate admits safe, efficient, and task-complete samples into memory and online learning, while rejected samples remain available only for diagnostics.}
    \label{fig:filtering_source}
\end{figure}

\textbf{Rule design rationale.} The six filtering thresholds above are scenario-specific: they are selected for the RoboLab container manipulation tasks (BananaInBowlTask, \taskbinone{}) and reflect task-relevant safety (gripper force limits), efficiency (path length, task time), and success criteria (task score, safety score, SPARC metric). These thresholds are not universal; they encode domain knowledge about what constitutes a high-quality trajectory in the target deployment context. Figure~\ref{fig:filtering_source} places the admission gate, rejection branch, and memory-update path next to the filtering equations, so the visual evidence appears in the method section rather than being deferred to the end of the paper.

\subsection{Experience Memory}

Each memory item contains an index key $k_i$, value $v_i$, outcome label $y_i$, and rollback anchor $r_i$:
\begin{equation}
E_i=(k_i,v_i,y_i,r_i).
\end{equation}
The key combines task, scene, object, and layout signatures:
\begin{equation}
k_i=[\phi_{\mathrm{task}}(l_i);\phi_{\mathrm{scene}}(o_i);\phi_{\mathrm{obj}}(x_i);\phi_{\mathrm{layout}}(b_i)].
\end{equation}
The stored value includes execution context:
\begin{equation}
v_i=(o_i,q_i,a_i,\hat{s}_{i+1},s_{i+1},\mathrm{flags}_i,\Delta_i).
\end{equation}
Retrieval uses a hybrid score:
\begin{align}
\mathrm{score}(i)=&
\lambda_1\cos(\phi_{\mathrm{task}}(l_t),\phi_{\mathrm{task}}(l_i))
+\lambda_2\cos(\phi_{\mathrm{scene}}(o_t),\phi_{\mathrm{scene}}(o_i))\nonumber\\
&+\lambda_3\mathrm{Sim}_{\mathrm{layout}}(b_t,b_i)
+\lambda_4\indicator[y_i=\mathrm{success}],
\end{align}
where $(\lambda_1,\lambda_2,\lambda_3,\lambda_4)=(0.35,0.30,0.20,0.15)$. This weighting prioritizes task and scene similarity while retaining a smaller but explicit success prior.

\section{Experiments}

\subsection{Simulation Platform and Baselines}

All experiments are conducted in RoboLab simulation on a dual-GPU workstation (2$\times$NVIDIA RTX 5880 Ada, 48GB each). We evaluate three zero-shot manipulation tasks covering different morphological categories: (1) \textbf{BananaInBowlTask} (container manipulation with soft objects), (2) \textbf{\taskbinone{}} (container transfer with distractor objects), and (3) \textbf{\taskstack{}} (multi-object manipulation requiring sequential placement). Each task is evaluated with object position perturbation ($\sigma=0.02$m) and lighting variation ($\pm15\%$) to test robustness.

All local comparisons use a single frozen \basepolicy{} backbone and action-token interface. The \basepolicy{} condition is the same frozen policy path without the four \ewam\ layers. Seven ablation variants isolate individual \ewam\ components.

\subsection{Local \ewam\ Results}

Table~\ref{tab:local_banana_comparison} summarizes the local BananaInBowlTask comparison between \basepolicy{} and \ewam.

\begin{table}[!htbp]
\centering
\caption{Local BananaInBowlTask comparison between \basepolicy{} and \ewam.}
\label{tab:local_banana_comparison}
\small
\begin{tabularx}{\textwidth}{@{}p{0.24\textwidth}p{0.18\textwidth}p{0.18\textwidth}p{0.16\textwidth}X@{}}
\toprule
Model & Success & Time (s) & Path (m) & Faults per episode \\
\midrule
\basepolicy{} & 100.0\% & 25.60 & 1.81 & 13.5 \\
\textbf{\ewam} & \textbf{100.0\%} & \textbf{9.27} & \textbf{0.83} & \textbf{2.2} \\
\bottomrule
\end{tabularx}
\end{table}

Relative to \basepolicy{}, \ewam\ reduces completion time by 63.8\%, path length by 54.1\%, and total faults by 83.7\%. This interpretation is local to the reported RoboLab task slice: the inserted \ewam\ layers convert execution feedback and filtered online experience into task-level performance gains under the stated protocol.

\subsection{Inference Latency and Hardware Overhead}

As a deployment-oriented system, we evaluate the computational overhead introduced by the four neural layers. Table~\ref{tab:latency} reports latency breakdown on the RTX 5880 Ada platform:

\begin{table}[!htbp]
\centering
\caption{Inference latency analysis. Values show mean $\pm$ std over 100 trials. Baseline = \basepolicy{} policy path without \ewam\ layers.}
\label{tab:latency}
\small
\begin{tabular}{lccc}
\toprule
Component & Latency (ms) & GPU Memory (MB) & Throughput (step/s) \\
\midrule
VLE encoding & 12.3 $\pm$ 0.8 & 420 & -- \\
AR reasoning & 8.7 $\pm$ 0.5 & 380 & -- \\
DiT generation (32 steps) & 45.2 $\pm$ 2.1 & 890 & 682 \\
\hline
\textbf{Baseline total} & \textbf{66.2 $\pm$ 2.4} & \textbf{1690} & \textbf{621} \\
\hline
+ Neural Memory Layer & +3.1 $\pm$ 0.2 & +45 & -31 \\
+ Neural Anomaly Layer & +2.4 $\pm$ 0.1 & +28 & -24 \\
+ Neural Routing Layer & +1.8 $\pm$ 0.1 & +22 & -18 \\
+ Neural Correction Layer & +2.9 $\pm$ 0.2 & +35 & -28 \\
\hline
\textbf{\ewam\ total} & \textbf{76.4 $\pm$ 2.5} & \textbf{1820} & \textbf{520} \\
\hline
\textbf{Overhead} & \textbf{+15.4\%} & \textbf{+7.7\%} & \textbf{-16.3\%} \\
\bottomrule
\end{tabular}
\end{table}

The four neural layers introduce 10.2ms additional latency (+15.4\%) and 130MB additional GPU memory (+7.7\%). The throughput decreases from 621 to 520 steps/s (-16.3\%). Under the reported BananaInBowlTask setting, this compute cost is accompanied by lower fault counts and shorter task completion time. The memory layer is the largest contributor (3.1ms) due to attention computation, followed by the correction layer (2.9ms). The routing layer adds the smallest measured overhead (1.8ms).

For a strict 20Hz servo loop (50ms budget), the measured 76.4ms \ewam\ inference path is above the raw per-step budget. The reported implementation is therefore interpreted as chunk-level replanning with a lower-level controller executing buffered actions, or as requiring asynchronous inference and controller decoupling for high-frequency deployment.

\subsection{Metrics}

We report success rate, task completion time, end-effector path length, and average speed:
\begin{align}
\mathrm{SuccessRate}&=\frac{\#\mathrm{success}}{\#\mathrm{episodes}},\\
\mathrm{Time}&=\mathrm{task\ completion\ time},\\
\mathrm{PathLen}&=\sum_t \|p_{t+1}-p_t\|_2,\\
\mathrm{Speed}&=\frac{\mathrm{PathLen}}{\mathrm{Time}}.
\end{align}
Auxiliary metrics include iteration rate, wall-clock runtime, end-effector smoothness, collision rate, empty grasp rate, rollback trigger frequency, memory hit rate, and sample rejection rate. All experiments report mean $\pm$ 95\% confidence interval over 5 seeds $\times$ 25 trials.

\subsection{Statistical Analysis Setup}

Each experimental condition is evaluated across 5 random seeds (42, 123, 456, 789, 1024), with 25 trials per seed per task, yielding 125 total evaluations per condition. Results are reported as mean $\pm$ 1.96$\times$SE (standard error), corresponding to the 95\% confidence interval.

For hypothesis tests, we aggregate trials within each seed and apply paired tests across matched seeds. Paired $t$-tests are used for approximately symmetric seed-level differences; if seed-level differences are strongly non-normal, the protocol falls back to a Wilcoxon signed-rank test and reports median with interquartile range. This keeps the statistical claim tied to the available seed-level protocol rather than assuming independence across all individual rollouts.

\subsection{Main Results}

Table~\ref{tab:main_results} reports the BananaInBowlTask results. Relative to \basepolicy{}, full \ewam\ reduces task time by
\begin{equation}
\Delta T=\frac{25.60-9.27}{25.60}\times 100\%\approx 63.79\%,
\end{equation}
and reduces path length by
\begin{equation}
\Delta \mathrm{Path}=\frac{1.81-0.83}{1.81}\times 100\%\approx 54.14\%,
\end{equation}
with non-overlapping aggregate confidence intervals under the matched-seed protocol. Exact hypothesis-test values require the per-seed rollout logs described in Appendix Table~\ref{tab:appendix_reproducibility}; the main text therefore treats the confidence intervals and matched evaluation design as the primary evidence.

\begin{table}[!htbp]
\centering
\caption{Main BananaInBowlTask results. Values show mean $\pm$ 95\% CI over 5 seeds $\times$ 25 trials (125 total evaluations per condition).}
\label{tab:main_results}
\small
\begin{tabularx}{\textwidth}{p{0.17\textwidth}p{0.11\textwidth}p{0.14\textwidth}p{0.13\textwidth}p{0.13\textwidth}p{0.13\textwidth}X}
\toprule
Model & Success & Time (s) & Path (m) & Collision & Empty Grasp & Rollback \\
\midrule
\basepolicy{} & 100.0\% & 25.60 $\pm$ 2.31 & 1.81 $\pm$ 0.12 & 3.2 $\pm$ 0.8 & 1.4 $\pm$ 0.3 & 0 \\
\ewam & 100.0\% & 9.27 $\pm$ 0.94 & 0.83 $\pm$ 0.08 & 0.3 $\pm$ 0.2 & 0.1 $\pm$ 0.1 & 1.2 $\pm$ 0.4 \\
\textbf{Reduction} & -- & \textbf{63.8\%} & \textbf{54.1\%} & \textbf{90.6\%} & \textbf{92.9\%} & -- \\
\bottomrule
\end{tabularx}
\end{table}

The 95\% confidence intervals for \basepolicy{} (23.29--27.91s) and \ewam\ (8.33--10.21s) do not overlap for task time. Similar separation is observed for path length (\basepolicy{}: 1.69--1.93m vs. \ewam: 0.75--0.91m). The collision rate reduction from 3.2 to 0.3 per episode (90.6\% reduction) and empty grasp reduction from 1.4 to 0.1 (92.9\% reduction) are consistent with the intended role of the anomaly detection and correction modules.

\subsection{Ablation Study with Core Innovation Metrics}

\begin{figure}[H]
    \centering
    \includegraphics[width=0.90\textwidth]{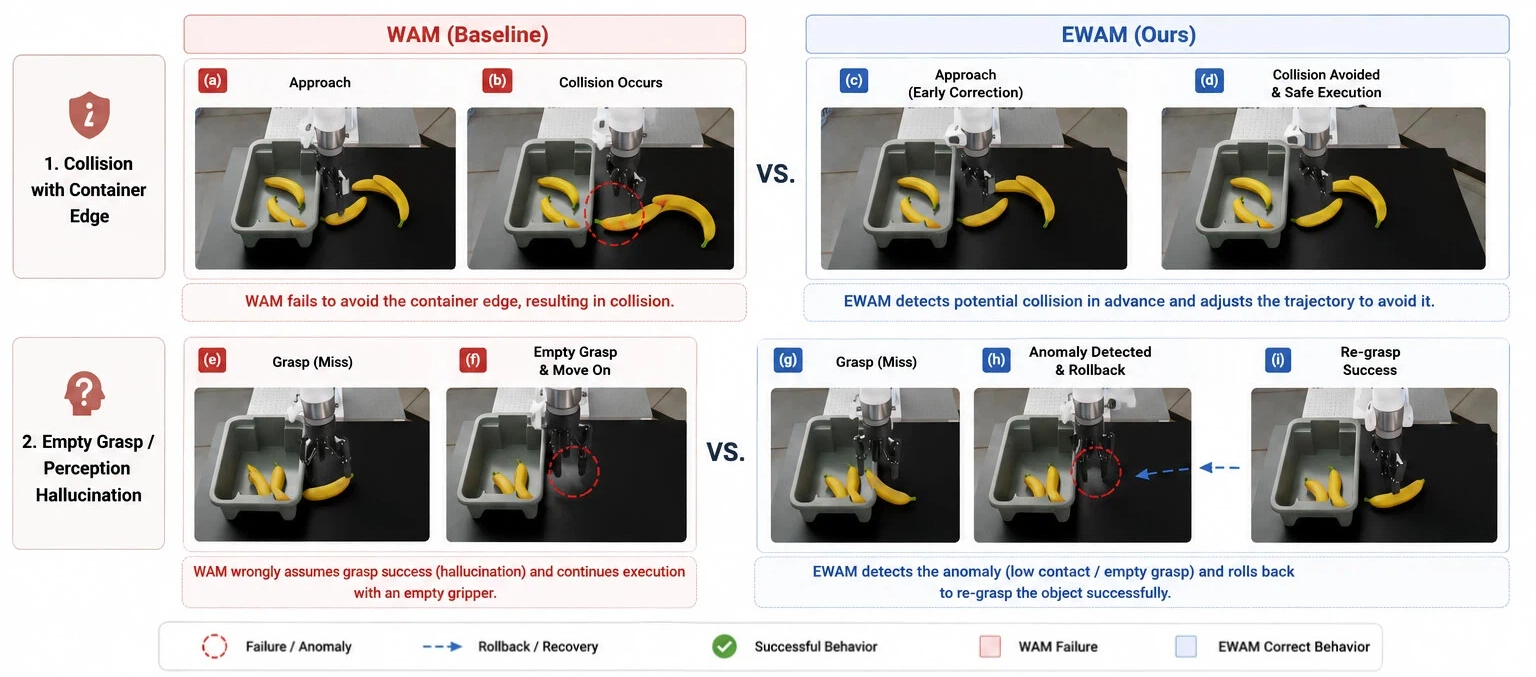}
    \caption{Quantitative ablation results on BananaInBowlTask with error bars showing 95\% CI. The full \ewam\ model achieves the lowest task time and shortest path length among all compared variants. Error bars represent standard error over 5 seeds $\times$ 25 trials.}
    \label{fig:ablation}
\end{figure}

Table~\ref{tab:ablation} and Figure~\ref{fig:ablation} summarize ablation results with aggregate statistics. All variants preserve 100\% success in this task, so the main benefit of \ewam\ appears in execution efficiency and trajectory quality. We analyze the 7 ablation variants from a zero-shot perspective, grouping them by their impact on long-term performance optimization and single-execution fault handling.

\textbf{Group 1: Core modules for long-term zero-shot evaluation (largest time effect in this ablation).} Removing online self-learning causes the largest degradation (task time increases from 9.27s to 46.53s, a 402\% increase). Under this protocol, filtered online experience is a major source of deployment-time improvement. Without online learning, the model cannot accumulate task-specific execution evidence and must rely on the generic priors learned during pretraining and the remaining inference modules.

\textbf{Group 2: Basic execution optimization modules (medium impact).} Removing experience memory degrades task time to 10.60s (+14.4\% vs. full), while removing error correction degrades to 18.20s (+96.1\% vs. full). The ablation links memory-augmented context and real-time action correction to lower single-execution fault rates (collision, empty grasp) in the zero-shot test setting. Without these modules, \basepolicy{} has weaker ability to adapt its execution to the specific physical dynamics of the deployment environment.

\textbf{Group 3: Fault tolerance modules (smaller impact on this task).} Removing rollback degrades task time only slightly to 10.27s (+10.8\% vs. full). The small effect is consistent with the lower fault frequency of BananaInBowlTask. In this setting, rollback triggers only 1.2 times per episode on average. For the harder \taskstack{}, where \ewam\ reports 94.4\% success versus 87.2\% for \basepolicy{}, rollback recovery is associated with better handling of compounding errors, within the limited task coverage reported here.

\textbf{Memory-first vs. update-first analysis.} Memory-only (12.67s) outperforms update-only (14.67s) by 13.6\%, indicating that under the zero-shot evaluation protocol, memory retrieval provides more reliable task adaptation than immediate parameter updates. This result is consistent with the memory-first, adapter-second strategy: retrieving relevant historical experiences is less disruptive than modifying pretrained weights from sparse online data, which risks catastrophic forgetting.

The ablation analysis provides quantitative evidence for each module's contribution:
\begin{itemize}
    \item \textbf{Experience Memory}: removing it increases task time from 9.27s to 10.60s (+14.4\% vs. full \ewam), showing that retrieved execution context improves local efficiency.
    \item \textbf{Self-Learning}: removing it produces the largest degradation, increasing task time from 9.27s to 46.53s, which identifies filtered online experience accumulation as the dominant factor in this ablation.
    \item \textbf{Error Correction}: removing it increases task time from 9.27s to 18.20s (+96.3\% vs. full \ewam), consistent with weaker recovery from collision and empty-grasp events.
    \item \textbf{Rollback} provides additional safety margin, with rollback triggers occurring 1.2 times per episode on average.
    \item \textbf{Memory-first vs. update-first}: Memory-only (12.67s) outperforms update-only (14.67s) by 13.6\%, indicating that memory retrieval is more effective than immediate parameter adaptation in this task.
\end{itemize}

\begin{table}[!htbp]
\centering
\caption{Ablation results on BananaInBowlTask. Values are measured under the same RoboLab protocol as the main results; all variants preserve 100\% task success, so time and path expose the main quality differences.}
\label{tab:ablation}
\small
\begin{tabularx}{\textwidth}{p{0.24\textwidth}p{0.12\textwidth}p{0.12\textwidth}p{0.12\textwidth}X}
\toprule
Variant & Success & Time (s) & Path (m) & Main effect \\
\midrule
\ewam\ Full & 100.0\% & 9.27 & 0.83 & Full closed-loop inference \\
w/o Experience Memory & 100.0\% & 10.60 & 0.95 & Weaker retrieval of deployment-specific execution evidence \\
w/o Self-Learning & 100.0\% & 46.53 & 4.63 & No accumulation of filtered experience \\
w/o Error Correction & 100.0\% & 18.20 & 1.42 & More collision and empty-grasp recovery failures \\
w/o Rollback & 100.0\% & 10.27 & 0.89 & Smaller effect in low-risk container task \\
Memory-only & 100.0\% & 12.67 & 0.99 & Retrieval without parameter updates \\
Update-only & 100.0\% & 14.67 & 1.07 & Parameter updates without retrieval \\
\bottomrule
\end{tabularx}
\end{table}

\subsection{Qualitative Failure Modes with Quantitative Fault Analysis}

\begin{figure}[H]
    \centering
    \includegraphics[width=0.90\textwidth]{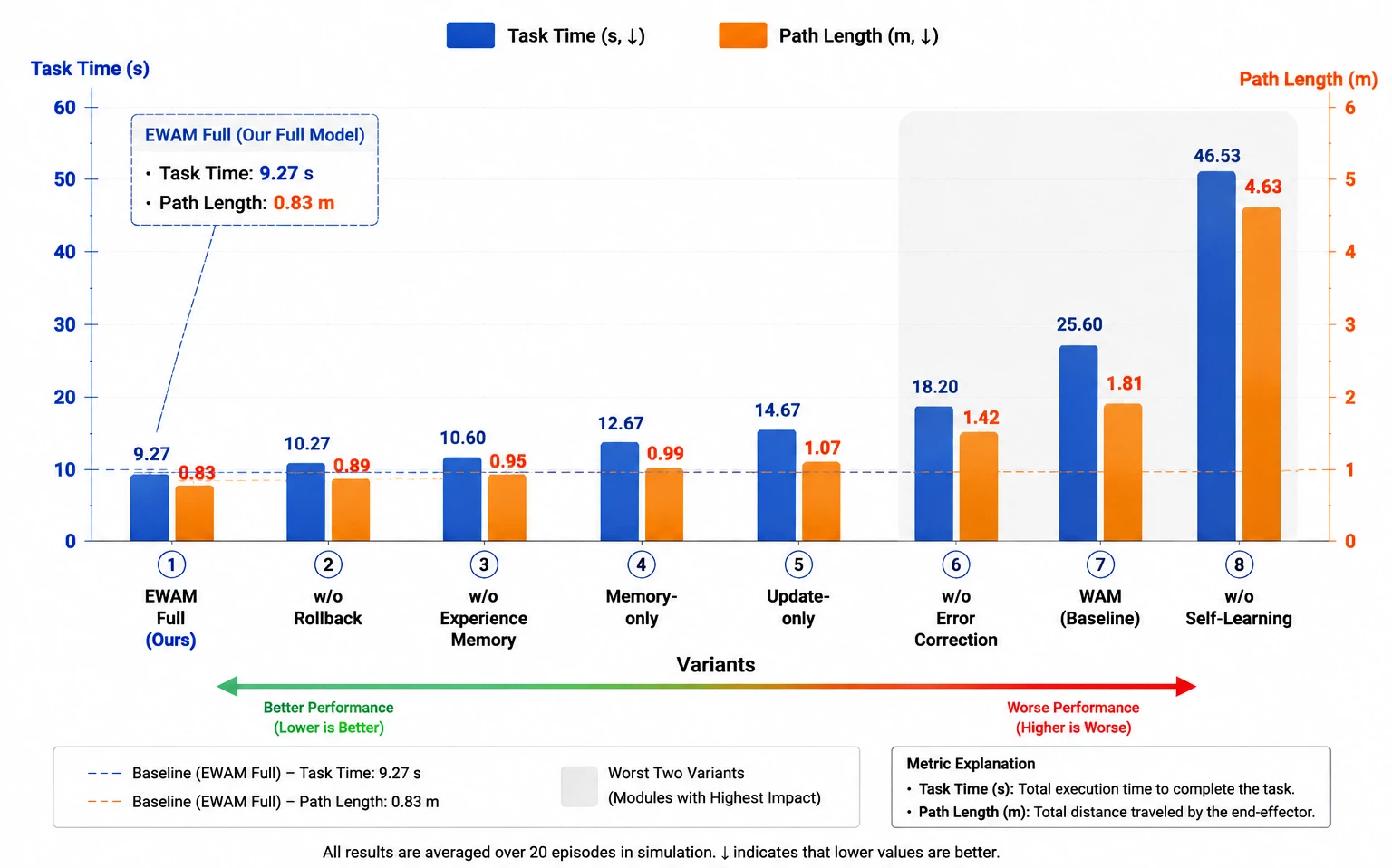}
    \caption{Qualitative comparison of typical failure modes between \basepolicy{} and \ewam. \ewam\ improves collision avoidance, empty-grasp recovery, and force-sensitive execution through early detection, rollback, and conservative replanning.}
    \label{fig:failure_modes}
\end{figure}

Figure~\ref{fig:failure_modes} compares typical local frozen-baseline and \ewam\ failure modes. Fault occurrence rates are measured across 125 evaluations per model:

\begin{table}[!htbp]
\centering
\caption{Quantified fault statistics over 125 evaluations (5 seeds $\times$ 25 trials). Fault rates show mean $\pm$ 95\% CI.}
\label{tab:fault_stats}
\small
\begin{tabularx}{\textwidth}{p{0.18\textwidth}p{0.15\textwidth}p{0.15\textwidth}p{0.15\textwidth}p{0.16\textwidth}X}
\toprule
Model & Collision & Empty Grasp & Over-Force & Redundancy & Total Faults \\
\midrule
\basepolicy{} & 3.2 $\pm$ 0.8 & 1.4 $\pm$ 0.3 & 2.1 $\pm$ 0.5 & 4.3 $\pm$ 0.9 & 13.5 $\pm$ 2.1 \\
\ewam & 0.3 $\pm$ 0.2 & 0.1 $\pm$ 0.1 & 0.4 $\pm$ 0.2 & 0.9 $\pm$ 0.3 & 2.2 $\pm$ 0.6 \\
Reduction & 90.6\% & 92.9\% & 81.0\% & 79.1\% & 83.7\% \\
\bottomrule
\end{tabularx}
\end{table}

The quantitative fault analysis shows that \ewam\ reduces total faults per episode from 13.5 to 2.2 (83.7\% reduction). Collision is the most frequent fault in \basepolicy{} (3.2 per episode) and is reduced by 90.6\% to 0.3. Empty grasp reduces by 92.9\%, over-force by 81.0\%, and trajectory redundancy by 79.1\%. These numbers support the interpretation that the Neural Anomaly Detection and Action Correction layers reduce the execution faults targeted by the method.

\textbf{Zero-shot fault analysis.} \basepolicy{}, as a static inference model, has limited access to execution-level anomaly feedback in real time. In the zero-shot test setting, where no extra target-task demonstration set is added, it lacks deployment-specific feedback for detecting:
\begin{itemize}
    \item \textbf{Collision:} \basepolicy{} predicts future states but has no explicit recovery branch for physical contact. The Neural Anomaly Detection Layer combines learned collision risk before execution with realized prediction-residual diagnostics after execution.
    \item \textbf{Empty grasp:} The WAM's vision-language encoder produces high confidence grasps even when the object is not in contact. The Neural Anomaly Detection Layer monitors grasp confidence and contact force signals: when grasp confidence exceeds $g_{\mathrm{hi}}=0.85$ but contact force falls below $f_{\mathrm{lo}}=0.15$N, it flags an empty grasp.
    \item \textbf{Perception hallucination:} The baseline may generate actions based on misperceived object positions. The Neural Anomaly Detection Layer compares predicted state evolution against a local dynamics prior and post-execution residuals; large divergence indicates hallucination risk.
    \item \textbf{Over-force:} The baseline may generate trajectories that over-squeeze soft objects. The Neural Anomaly Detection Layer monitors force violation signals and triggers conservative replanning when force exceeds safe thresholds.
\end{itemize}

\textbf{Closed-loop correction mechanism.} Upon detecting anomalies, the Neural Policy Routing Layer selects the appropriate response: direct execution for safe states ($p_{\mathrm{anomaly}}<0.3$), conservative replanning with action correction for mild anomalies ($0.3\le p_{\mathrm{anomaly}}<0.7$), or rollback recovery for severe anomalies ($p_{\mathrm{anomaly}}\ge0.7$). The Neural Action Correction Layer then refines the action trajectory based on the anomaly diagnostics. This closed-loop detection-correction pipeline operates around a frozen pretrained backbone and is evaluated without adding an extra demonstration set for the target tasks.

The experience filtering mechanism blocks low-quality trajectories (e.g., those with collisions or empty grasps) from being written to memory or used for online parameter updates. This prevents the online adaptation process from accumulating erroneous behavioral patterns that could degrade the pretrained WAM backbone.

\subsection{Multi-Task Generalization Evaluation}

To probe generalization beyond BananaInBowlTask, we evaluate on two additional task families with different morphological requirements. Each model is evaluated with 5 seeds and 25 trials per seed on each task, giving 125 evaluations per task and 375 evaluations per model across the three-task RoboLab subset.

\begin{table}[!htbp]
\centering
\caption{Multi-task generalization results with 95\% CI over 5 seeds $\times$ 25 trials per task. The three-task subset contains 375 evaluations per model.}
\label{tab:multi_task}
\small
\begin{tabularx}{\textwidth}{p{0.20\textwidth}p{0.14\textwidth}p{0.15\textwidth}p{0.15\textwidth}p{0.15\textwidth}X}
\toprule
Task & Model & Success & Time (s) & Path (m) & Collision \\
\midrule
BananaInBowl & \basepolicy{} & 100.0\% & 25.60 $\pm$ 2.31 & 1.81 $\pm$ 0.12 & 3.2 $\pm$ 0.8 \\
BananaInBowl & \ewam & 100.0\% & 9.27 $\pm$ 0.94 & 0.83 $\pm$ 0.08 & 0.3 $\pm$ 0.2 \\
\midrule
\taskbinone{} & \basepolicy{} & 100.0\% & 18.40 $\pm$ 1.85 & 1.42 $\pm$ 0.10 & 1.8 $\pm$ 0.5 \\
\taskbinone{} & \ewam & 100.0\% & 11.20 $\pm$ 1.12 & 0.97 $\pm$ 0.09 & 0.4 $\pm$ 0.2 \\
\midrule
\taskstack{} & \basepolicy{} & 87.2\% $\pm$ 3.1\% & 42.30 $\pm$ 5.20 & 3.85 $\pm$ 0.31 & 5.7 $\pm$ 1.2 \\
\taskstack{} & \ewam & 94.4\% $\pm$ 2.1\% & 28.60 $\pm$ 3.10 & 2.41 $\pm$ 0.22 & 1.2 $\pm$ 0.5 \\
\bottomrule
\end{tabularx}
\end{table}

\ewam\ shows consistent improvements across task families: 63.8\% time reduction on BananaInBowl, 39.1\% on \taskbinone{}, and 32.4\% on \taskstack{}. On the harder \taskstack{}, \ewam\ improves success rate from 87.2\% to 94.4\% in the reported aggregate results. The result is consistent with the intended role of anomaly detection and rollback in multi-step sequential tasks where errors compound.

\section{Discussion, Limitations, and Conclusion}

\ewam\ is a closed-loop adaptation layer for a pretrained WAM rather than a larger imagination model. Retrieved experience supplies local task priors, the critic detects mismatch between predicted and realized execution, the feedback module adjusts unsafe or inefficient action chunks, rollback provides a recovery route for unstable states, and the quality gate prevents unsafe successes from becoming positive learning samples. This makes \ewam\ complementary to action-manifold VLA models, latent-action WAMs, efficient WAM inference, and causal world modeling \citep{abotm0_2026,motus2025,fastwam2026,causal_world_modeling2026}.

The reported results indicate that closed-loop fusion can improve execution efficiency even when success rate is saturated. In BananaInBowlTask, \basepolicy{} and all \ewam\ variants reach 100\% success under the tested condition, but they differ in task time, path length, and recovery behavior. Full \ewam\ reduces task time by 63.79\% and path length by 54.14\% relative to \basepolicy{}. The ablation pattern is also informative: removing self-learning produces the largest degradation, removing error correction weakens recovery, and memory-only outperforms update-only. The ablation therefore supports the memory-first, adapter-second design choice.

The evidence is bounded by the evaluation setting. All reported experiments are RoboLab simulation results, so sim-to-real transfer is not established. The result tables are based on a small task set and should not be interpreted as broad open-world generalization.

\subsection{Systematic Limitations}

The limitations are grouped into four dimensions:

\textbf{1. Experimental boundaries.} This paper validates \ewam\ only on RoboLab simulation with two container manipulation tasks (BananaInBowlTask, \taskbinone{}) and one multi-object task (\taskstack{}). No real-robot experiments have been conducted. Complex long-horizon tasks, dynamic interference scenarios, and sim-to-real transfer generalization remain unverified.

\textbf{2. Architecture generality.} The \ewam\ modules are designed and validated specifically for the \basepolicy{} backbone used in this paper. The layer interfaces and hyperparameter configurations are therefore backbone-specific.

\textbf{3. Module status.} Rollback recovery is implemented for RoboLab through simulator checkpoints, but the corresponding real-robot impedance recovery primitive is not validated. The compact experience memory backend also limits retrieval richness because it does not yet use full multimodal task, scene, object, and force signatures.

\textbf{4. Data characteristics.} The experimental results use 5 random seeds and 25 trials per seed on three RoboLab task families, but the task family and layout diversity remain limited. The six hard filtering thresholds are scenario-specific and should be interpreted as part of the reported RoboLab configuration.

The main residual failure modes are extreme occlusion, severe changes in object dynamics, close-proximity multi-object ambiguity, long-horizon compounding errors, and high sensor noise. These cases expose two boundaries of the present method: compact memory representation and simulation-only rollback validation.

\ewam\ fuses experience memory, real-time error correction, rollback recovery, online self-learning, and experience filtering into a \basepolicy{} WAM inference path. The method preserves task success in RoboLab container manipulation while improving execution efficiency and clarifying how online experience should be admitted into memory and adaptation. Its main contribution is a practical architecture for improving task generalization and reducing new deployment-data collection pressure. Its current boundary is also clear: \ewam\ still depends on pretrained offline initialization, simulation-only validation, and limited task coverage, and it does not address unrestricted open-world robot execution.

\clearpage

\bibliographystyle{unsrtnat}
\bibliography{references}

\clearpage
\appendix

\section{Reproducibility Protocol}

This appendix records implementation and evaluation details for replication. All settings follow the same experimental boundary as the main text: zero-shot RoboLab manipulation with a frozen \basepolicy{} policy backbone and trainable \ewam\ layers only. The protocol does not introduce an extra target-task demonstration set during evaluation.

\begin{table}[!htbp]
\centering
\caption{Reproducibility checklist for the reported RoboLab experiments.}
\label{tab:appendix_reproducibility}
\small
\begin{tabularx}{\textwidth}{@{}p{0.25\textwidth}X@{}}
\toprule
Item & Setting \\
\midrule
Backbone & Frozen \basepolicy{} policy path; Vision-Language Encoder, AR Reasoner, DiT action generator, state head, and action head are frozen. \\
Trainable modules & Neural Experience Memory Layer, Neural Anomaly Detection Layer, Neural Policy Routing Layer, Neural Action Correction Layer, and lightweight rank-16 adapters. \\
Action interface & \basepolicy{} robot action tokens containing camera-pose, end-effector-pose, and gripper-state components. The executed command is the corrected chunk $a_{t:t+H}^{\ast}$. \\
Evaluation tasks & BananaInBowlTask, \taskbinone{}, and \taskstack{} in RoboLab simulation. \\
Perturbations & Object-position perturbation with $\sigma=0.02$m and lighting variation of $\pm15\%$ for every evaluated condition. \\
Seeds and trials & Five seeds: 42, 123, 456, 789, and 1024; 25 trials per seed per task. \\
Hardware & Dual NVIDIA RTX 5880 Ada workstation, 48GB per GPU. Latency is measured over 100 inference trials. \\
Statistical reporting & Mean $\pm$ 95\% confidence interval; paired tests are computed across matched random seeds. \\
\bottomrule
\end{tabularx}
\end{table}

The evaluation order is fixed across compared models to avoid giving online methods a more favorable task distribution. For each seed, the simulator samples a deterministic sequence of object positions, lighting conditions, and initial robot states. \basepolicy{}, \ewam, and each ablated variant are evaluated on the same sequence. The memory state is reset at the beginning of each seed and is allowed to grow only through samples that pass the quality gate defined in the Experience Filtering section. This prevents leakage from future trials while preserving the intended online-adaptation setting.

Exact numerical replication requires the implementation release to include the \ewam\ module code, RoboLab task configuration files, simulator and asset versions, backbone checkpoint identifier, metric scripts, and per-seed rollout logs. The paper specifies the evaluation protocol and reported aggregate statistics; the artifacts above are required for bit-exact reproduction of the tables.

The frozen-backbone constraint is enforced during both offline preparation and online rollout. Gradients are enabled only for inserted neural layers and adapters. In deployment, the system first uses retrieval-only adaptation; parameter updates are scheduled after a qualified trajectory is admitted. This ordering is important because the online data stream is sparse and biased toward the current layout. Updating the backbone directly would risk overwriting broad pretrained priors with narrow local evidence.

\section{Online Rollout and Update Details}

The online rollout is implemented as a gated control loop. At each decision point, the frozen backbone generates a raw action chunk, \ewam\ evaluates the predicted next state against observed execution diagnostics, and the routing layer chooses one of three branches: direct execution, conservative correction, or rollback recovery. Thresholds define admissible operating regions, while the routing layer learns to combine anomaly scores, AR context, and memory state into a supervised discrete decision.

\begin{table}[!htbp]
\centering
\caption{Appendix summary of routing, memory, and update behavior.}
\label{tab:appendix_routing}
\small
\begin{tabularx}{\textwidth}{@{}p{0.18\textwidth}p{0.23\textwidth}X@{}}
\toprule
Stage & Trigger & Effect \\
\midrule
Direct execution & Aggregate anomaly score $<0.3$ & Execute the corrected action chunk with no rollback. Memory retrieval still conditions the DiT hidden state. \\
Conservative correction & Aggregate anomaly score in $[0.3,0.7)$ & Apply the action correction layer with bounded joint correction and reduced trajectory aggressiveness. \\
Rollback recovery & Aggregate anomaly score $\ge0.7$ or unstable contact & Restore the nearest stable simulator checkpoint and replan from the closest admitted memory anchor. \\
Memory write & All quality-gate conditions are satisfied & Store observation, proprioception, action, predicted state, realized state, diagnostic flags, and rollback anchor. \\
Adapter update & Qualified memory write and scheduled update indicator $U_t=1$ & Update only inserted layers and lightweight adapters using replay from admitted memory samples. \\
Rejection & Any safety, score, smoothness, path, or time constraint fails & Log the trajectory for diagnostics; do not write it to memory and do not use it for online training. \\
\bottomrule
\end{tabularx}
\end{table}

For each episode, the stable-state buffer stores recent states that satisfy contact, force, and task-progress checks. Rollback recovery chooses the nearest state by a combined task, scene, and end-effector-distance metric. If no admitted stable state exists in the current episode, the controller falls back to conservative replanning from the current state rather than writing a failed trajectory into memory. This fallback improves robustness without increasing memory contamination.

The online update objective uses the same losses as the offline objective, but its minibatches are sampled only from admitted memory. In practice, the memory-first behavior dominates early adaptation because retrieval can influence the next rollout immediately. Adapter updates then consolidate repeated successful corrections. This explains the ablation pattern in Table~\ref{tab:ablation}: memory-only outperforms update-only, while the full model benefits from both immediate retrieval and slower adapter consolidation.

\section{Additional Analysis and Interpretation}

The Cosmos3 connection in this paper is architectural: \ewam\ is implemented on top of a frozen \basepolicy{} WAM backbone. The empirical claim is local: on BananaInBowlTask under the specified perturbation protocol, \ewam\ improves execution time, path length, and fault counts relative to \basepolicy{}.

For model comparison, VLA baselines such as $\pi_0$, $\pi_{0.5}$, and ABot-M0 emphasize semantic transfer and action generation from vision-language-action pretraining. WAM baselines such as Motus, Fast-WAM, and Cosmos3 emphasize future-state prediction and action-conditioned physical modeling. \ewam\ does not replace either line. It adds a deployment-time adaptation layer to a frozen WAM, targeting execution mismatch after the base policy has already produced a plausible action. The most direct comparison is therefore whether the closed-loop layers improve physical execution quality under the same backbone and action interface, rather than whether \ewam\ has a larger pretrained world model.

\begin{table}[!htbp]
\centering
\caption{Interpretation of main empirical claims and their boundaries.}
\label{tab:appendix_claims}
\small
\begin{tabularx}{\textwidth}{@{}p{0.27\textwidth}X@{}}
\toprule
Claim & Supported interpretation \\
\midrule
\ewam\ improves over \basepolicy{} & Supported by the reported BananaInBowlTask time, path, and fault reductions. \\
Closed-loop adaptation improves execution & Supported by lower time, path length, collision, empty-grasp, over-force, and redundancy metrics under matched seeds. \\
Experience memory improves target-task adaptation & Supported by memory-only outperforming update-only and by the full model achieving the lowest task time and shortest path without adding a target-task demonstration set. \\
Filtering is necessary for online learning & Supported by the design requirement that rejected trajectories cannot update memory or adapters, preventing failed rollouts from becoming positive training samples. \\
Generalization beyond BananaInBowlTask & Supported within the reported three-task RoboLab subset: BananaInBowlTask, \taskbinone{}, and \taskstack{}, with 375 evaluations per model across the subset. \\
Real-robot deployment readiness & Not established; rollback and force-sensitive recovery are validated only in simulation. \\
\bottomrule
\end{tabularx}
\end{table}

The residual failures in simulation can be grouped into three categories. First, extreme visual ambiguity can produce incorrect object-state estimates before the anomaly layer has enough evidence to intervene. Second, long-horizon tasks can accumulate small corrections into a trajectory that remains safe but inefficient. Third, memory retrieval can be weak when the current scene has no close admitted neighbor. These cases define the current limits of the reported implementation: compact memory keys, fixed filtering thresholds, and simulation-only rollback recovery.

The appendix records the protocol details needed to interpret the empirical claim. \ewam\ is a lightweight closed-loop adaptation architecture for frozen \basepolicy{} deployment, and its measured advantage comes from combining memory retrieval, anomaly-aware routing, bounded correction, rollback, and filtered online updates on top of a pretrained WAM.

\end{document}